\documentclass[sigconf,nonacm]{acmart}

\renewcommand\footnotetextcopyrightpermission[1]{}

\acmConference[MLF @ KDD 2026]{9th Workshop on Machine Learning in Finance, co-located with KDD 2026}{August 9, 2026}{Jeju, Korea}

\usepackage{subcaption}
\usepackage{colortbl}
\usepackage{placeins}
\usepackage{booktabs}
\usepackage[utf8]{inputenc}
\usepackage{newunicodechar}
\usepackage{tikz}
\usetikzlibrary{arrows.meta,calc,positioning}
\usepackage{dblfloatfix}
\newunicodechar{≤}{\leq}
\usepackage{makecell}

\AtBeginDocument{%
  }

\begin{document}
\raggedbottom
\newcommand{\E}{\mathbb{E}}
\newcommand{\Var}{\mathrm{Var}}
\newcommand{\Law}{\mathcal{L}}
\newcommand{\sd}{\mathrm{sd}}

\counterwithout{theorem}{section}
\renewcommand{\thetheorem}{\arabic{theorem}}

\title{Deployment-Side Adaptiveness in Multi-Horizon Volatility Forecasting}
\titlenote{Accepted at the 9th Workshop on Machine Learning in Finance, co-located with ACM SIGKDD Conference on Knowledge Discovery and Data Mining (KDD 2026), Jeju, Korea, August 9, 2026.}

\author{Riku Green}
\email{riku.green@bristol.ac.uk}
\orcid{0009-0006-1215-1473}
\affiliation{%
  \institution{The University of Bristol}
  \city{Bristol}
  \country{UK}
}

\author{Zahraa S. Abdallah}
\email{zahraa.abdallah@bristol.ac.uk}
\affiliation{%
  \institution{The University of Bristol}
  \city{Bristol}
  \country{UK}
}

\author{Telmo M Silva Filho}
\email{telmo.silvafilho@bristol.ac.uk}
\affiliation{%
  \institution{The University of Bristol}
  \city{Bristol}
  \country{UK}
}

\begin{abstract}
In financial forecasting, predictive performance depends not only on which model is trained, but also on how the trained model is deployed. We study this issue in multi-horizon volatility forecasting. Our starting point is that a trained multi-output (MIMO) forecaster does not define a single deployable predictor: by changing the inference-time rollout rule, the same trained model induces a family of forecasts with different accuracy and cost profiles. Across 20 stock-volatility series, three forecast horizons, and architectures ranging from linear models to PatchTST, we find that non-default rollout rules often improve over standard MIMO deployment. However, the best fixed rule varies substantially across architectures and horizons, making any single static replacement unreliable. We therefore evaluate validation-based deployment policies over the induced rule family. Under the primary MSE objective, validation-selected singletons provide a low-cost improvement over default MIMO, while small rule subsets recover much of the benefit of larger ensembles at substantially lower inference cost. We also find that policy rankings are metric-sensitive: MSE-selected policies do not transfer uniformly to QLIKE, a finance-standard volatility loss. These results show that inference-time deployment is a meaningful source of adaptiveness in financial forecasting, and that trained volatility forecasters should be evaluated not only by their architecture, but also by their deployment policy.
\end{abstract}

\keywords{time series forecasting, financial forecasting, ensemble methods, multistep forecasting, efficiency}

\maketitle

\section{Introduction}
\label{sec:intro}

Financial forecasting systems are deployed repeatedly across assets, horizons, and market conditions, often under latency, compute, and maintenance constraints. These issues are especially important in volatility forecasting, which underpins risk management \cite{christoffersen2000relevant}, volatility timing and allocation \cite{fleming2001economic}, and the realized-volatility pipeline used throughout empirical finance \cite{andersen2003modeling,bucci2017forecasting}. Recent evidence shows that machine learning methods can be competitive with strong classical volatility baselines \cite{christensen2023machine}, while surveys document the growing role of modern ML in financial forecasting \cite{wasserbacher2022machine,masini2023machine,behera2024machine,giantsidi2025deep}. Yet forecasting systems are still commonly evaluated as though training alone determines the predictor that will be deployed.

This view is too narrow for multi-horizon forecasting. In multi-step prediction, the forecast used in practice depends not only on learned parameters, but also on the inference-time rule used to generate the horizon. Prior work shows that different multi-step forecasting strategies can produce materially different errors across datasets, horizons, and model classes, with no single strategy uniformly best \cite{green2025stratify,green2024time}. Recursive composition can also change the effective prediction problem faced at deployment, rather than merely accumulating error \cite{green2025epistemic}. Thus, deployment choices can materially alter the predictor used in practice, even after training is fixed.

This is particularly relevant for multi-output forecasting. A trained multi-output (MIMO) model is usually treated as defining one natural deployed predictor: the default direct multi-horizon forecast \cite{green2025stratify}. We instead view the same trained model as inducing a family of predictors. By committing smaller output blocks, updating the input state, and reusing the same forecasting function, a fixed MIMO model can generate distinct multi-horizon forecasts without retraining. We therefore ask: once a multi-horizon MIMO forecaster has been trained, is the default MIMO forecast necessarily its best operational deployment?

This deployment-side perspective is especially natural in finance. Practitioners care not only about average benchmark error, but also about repeated inference cost, heterogeneity across assets and horizons, and whether the selected policy matches the operational loss. This distinction is especially relevant in volatility forecasting, where long-horizon targets are highly uncertain, squared-error objectives can favor accurate but distributionally conservative conditional-mean forecasts, and log-scale MSE and finance-standard losses such as QLIKE emphasize different aspects of forecast quality \cite{green2026expectations}. Deployment is therefore a budgeted and metric-sensitive evaluation problem.

We study this perspective in daily multi-step stock-volatility forecasting. Across 20 stock-volatility series, multiple forecast horizons, and architectures ranging from linear models to PatchTST, we evaluate the forecast family induced by alternative inference-time rollout rules. Non-default rollout rules often improve over default MIMO deployment, but the strongest fixed rule varies across architectures and horizons, making any single static replacement unreliable.

Validation-based deployment policies exploit this induced rule family under practical inference budgets. Under the primary MSE objective, validation-selected singletons provide a low-cost improvement over default MIMO, while small rule subsets recover much of the benefit of larger ensembles at substantially lower inference cost. However, policy rankings are metric-sensitive: MSE-selected policies do not transfer uniformly to QLIKE. Finally, compared with matched policies built from separately trained models, induced policies achieve near-parity predictive performance at substantially lower retraining cost.

The main implication is that inference-time deployment is a meaningful source of adaptiveness in financial forecasting. Our contributions are as follows:
\begin{itemize}
    \item We introduce the induced-rule view of MIMO deployment for volatility forecasting: a single trained multi-output forecaster defines a family of deployable predictors through inference-time rollout choices.

    \item We show that this induced rule family provides useful MSE-oriented deployment headroom. Non-default rollout rules often improve over default MIMO, but the best fixed rule is architecture- and horizon-dependent; validation-selected singletons and small rule subsets exploit this headroom under practical inference budgets.

    \item We show that deployment choice is metric-sensitive in volatility forecasting. Policies selected under validation MSE do not transfer uniformly to QLIKE, so the preferred deployment policy depends on the operational loss.

    \item We compare induced-forecast policies with matched policies built from separately trained models, showing that induced policies can approach trained-policy accuracy while reducing the retraining cost needed to reproduce a selected deployment policy.
\end{itemize}

\section{Related Work}
\label{sec:related}

\noindent\textbf{Multi-step forecasting strategies.}
Multi-step forecasting is commonly organized around recursive, direct, hybrid, and multi-output strategies  \cite{green2025stratify}. Strategy choice can materially affect forecast error, and in the multi-output setting it also affects inference cost: recursive-style deployment with committed output size $s$ requires approximately $\lceil H/s\rceil$ rollout calls, whereas default MIMO deployment predicts the full horizon in one call. 
Recent work shows that no single multi-step strategy is uniformly best across datasets and function classes \cite{green2025stratify}, and that recursive and direct behavior can differ because recursive composition changes the effective dynamics and uncertainty propagation of the learned predictor \cite{green2025epistemic,green2026exposure}. 
Our focus is different. We do not compare separately trained forecasting strategies as alternative model classes. Instead, we study how a \emph{single trained MIMO model} can be redeployed through alternative inference-time rollout rules.

\noindent\textbf{Forecast combination and ensembling.}
Forecast combination has a long history, from classical work showing that combining forecasts can reduce mean-squared error \cite{bates1969combination}, to modern reviews documenting a wide range of successful weighting and aggregation schemes \cite{wang2023forecast}. More generally, ensemble methods are effective when component predictors are individually useful but not perfectly correlated \cite{dietterich2000ensemble,breiman2001random}, and forecasting evidence continues to support combinations as a strong baseline \cite{makridakis2018m4}. Prior work also emphasizes that the value of combination depends on the uniqueness and dependence structure of the component forecasts \cite{thomson2019combining,wu2021ensemble}. In multi-step forecasting, dynamic strategy selection and combination can help when the best fixed strategy is not knowable \emph{a priori} \cite{green2024time}. Our setting differs from both conventional ensemble construction and stochastic test-time ensembling. Rather than training many independent predictors or sampling randomized model evaluations, as in MC dropout \cite{gal2016dropout}, we obtain a deterministic family of forecasts from a single trained model by varying only the inference-time rollout rule. Diversity therefore comes from different compositions of the same forecasting map and the self-induced states visited during deployment, while the trained parameters remain fixed.

\begin{figure*}[t]
\centering
\begin{tikzpicture}[
    x=1cm,y=1cm,
    >=Latex,
    slot/.style={circle,draw=black,minimum size=4.8mm,inner sep=0pt},
    used/.style={circle,draw=blue!70!black,fill=blue!12,minimum size=4.8mm,inner sep=0pt},
    future/.style={circle,draw=black!35,minimum size=4.8mm,inner sep=0pt},
    faded/.style={circle,draw=black!20,minimum size=4.8mm,inner sep=0pt},
    state/.style={draw,rounded corners=2pt,fill=gray!8,minimum width=1.3cm,minimum height=.62cm,font=\small},
    rowlab/.style={font=\small\bfseries,align=right},
    expl/.style={font=\scriptsize,align=center},
    funarr/.style={->,blue!70!black,line width=1pt},
    grayarr/.style={->,black!55,line width=.8pt},
    fuse/.style={circle,draw=blue!70!black,thick,minimum size=4.8mm,inner sep=0pt,font=\scriptsize,text=blue!70!black},
    flabel/.style={font=\scriptsize,text=blue!70!black,fill=white,inner sep=1.2pt}
]
\colorlet{accent}{blue!70!black}

\def\yA{0.0}
\def\yB{-1.95}
\def\yC{-4.75}

\def\yFuseB{-3.05}
\def\yStateB{-3.80}

\def\yFuseC{-5.95}
\def\yStateC{-6.70}

\def\xstate{2.75}

\def\xone{5.55}
\def\xtwo{7.35}
\def\xthree{9.15}
\def\xfour{10.95}
\def\xfive{12.75}
\def\xsix{14.55}
\def\xbound{13.70}

\node[expl] at (\xone,0.95)   {$t\!+\!1$};
\node[expl] at (\xtwo,0.95)   {$t\!+\!2$};
\node[expl] at (\xthree,0.95) {$t\!+\!3$};
\node[expl] at (\xfour,0.95)  {$t\!+\!4$};
\node[expl] at (\xfive,0.95)  {$t\!+\!5$};

\draw[densely dotted,thick] (\xbound,1.15) -- (\xbound,-7.05);
\node[expl,anchor=west] at (14.05,\yA+0.25) {Forecast horizon\\ $H=5$};

\node[rowlab,anchor=east] at (1.25,\yA) {Default MIMO\\$(s=5)$};

\node[state] (x0) at (\xstate,\yA) {$\mathbf{x}_t$};
\draw[funarr]
    (x0.east) -- node[midway,above,flabel] {$f_\theta$} (\xone-0.55,\yA);

\node[used] at (\xone,\yA) {};
\node[used] at (\xtwo,\yA) {};
\node[used] at (\xthree,\yA) {};
\node[used] at (\xfour,\yA) {};
\node[used] at (\xfive,\yA) {};

\draw[accent,line width=1pt] (\xone-0.24,\yA-0.37) -- (\xfive+0.24,\yA-0.37);
\node[expl] at ({(\xone+\xfive)/2},\yA-0.78) {single-shot forecast};

\node[rowlab,anchor=east] at (1.25,\yB) {Block-recursive\\$(s=2)$};

\node[state] (xb0)  at (\xstate,\yStateB) {$\mathbf{x}_t$};
\node[state,text=accent] (xh2b) at ({(\xone+\xtwo)/2},\yStateB) {$\hat{\mathbf{x}}_{t+2}$};
\node[state,text=accent] (xh4b) at ({(\xthree+\xfour)/2},\yStateB) {$\hat{\mathbf{x}}_{t+4}$};

\node[fuse] (fuseb1) at ({(\xone+\xtwo)/2},\yFuseB) {$\oplus$};
\node[fuse] (fuseb2) at ({(\xthree+\xfour)/2},\yFuseB) {$\oplus$};

\draw[funarr]
    (xb0.north east) to[out=28,in=200]
    node[pos=.56,above,sloped,flabel] {$f_\theta$}
    (\xone-0.45,\yB+0.02);

\node[used]   at (\xone,\yB) {};
\node[used]   at (\xtwo,\yB) {};
\node[future] at (\xthree,\yB) {};
\node[future] at (\xfour,\yB) {};
\node[future] at (\xfive,\yB) {};
\draw[accent,line width=1pt] (\xone-0.24,\yB-0.37) -- (\xtwo+0.24,\yB-0.37);

\node[expl,text=black!60] at ({(\xone+\xtwo)/2},\yB+0.58) {(truncate output past $t\!+\!2$)};

\draw[grayarr] (xb0.east) to[out=8,in=180] (fuseb1.west);
\draw[grayarr] ({(\xone+\xtwo)/2},\yB-0.37) -- (fuseb1.north);
\draw[grayarr] (fuseb1.south) -- (xh2b.north);

\draw[funarr]
    (xh2b.north east) to[out=28,in=200]
    node[pos=.56,above,sloped,flabel] {$f_\theta$}
    (\xthree-0.45,\yB+0.02);

\node[used]   at (\xthree,\yB) {};
\node[used]   at (\xfour,\yB) {};
\node[future] at (\xfive,\yB) {};
\node[faded]  at (\xsix,\yB) {};
\node[expl,text=black!45] at (15.25,\yB) {$\cdots$};
\draw[accent,line width=1pt] (\xthree-0.24,\yB-0.37) -- (\xfour+0.24,\yB-0.37);

\draw[grayarr] (xh2b.east) to[out=8,in=180] (fuseb2.west);
\draw[grayarr] ({(\xthree+\xfour)/2},\yB-0.37) -- (fuseb2.north);
\draw[grayarr] (fuseb2.south) -- (xh4b.north);

\draw[funarr]
    (xh4b.north east) to[out=28,in=200]
    node[pos=.56,above,sloped,flabel] {$f_\theta$}
    (\xfive-0.45,\yB+0.02);

\node[used]  at (\xfive,\yB) {};
\node[faded] at (\xsix,\yB) {};
\draw[accent,line width=1pt] (\xfive-0.24,\yB-0.37) -- (\xfive+0.24,\yB-0.37);

\node[expl,anchor=west] at (14.05,\yB-0.65) {Truncated past\\ desired horizon};

\node[rowlab,anchor=east] at (1.25,\yC) {Fully recursive\\$(s=1)$};

\node[state] (xc0)  at (\xstate,\yStateC) {$\mathbf{x}_t$};
\node[state,text=accent] (xh1c) at (\xone,\yStateC) {$\hat{\mathbf{x}}_{t+1}$};
\node[state,text=accent] (xh2c) at (\xtwo,\yStateC) {$\hat{\mathbf{x}}_{t+2}$};
\node[state,text=accent] (xh3c) at (\xthree,\yStateC) {$\hat{\mathbf{x}}_{t+3}$};
\node[state,text=accent] (xh4c) at (\xfour,\yStateC) {$\hat{\mathbf{x}}_{t+4}$};

\node[fuse] (fusec1) at (\xone,\yFuseC) {$\oplus$};
\node[fuse] (fusec2) at (\xtwo,\yFuseC) {$\oplus$};
\node[fuse] (fusec3) at (\xthree,\yFuseC) {$\oplus$};
\node[fuse] (fusec4) at (\xfour,\yFuseC) {$\oplus$};

\draw[funarr]
    (xc0.north east) to[out=28,in=200]
    node[pos=.56,above,sloped,flabel] {$f_\theta$}
    (\xone-0.45,\yC+0.02);

\node[used]   at (\xone,\yC) {};
\node[future] at (\xtwo,\yC) {};
\node[future] at (\xthree,\yC) {};
\node[future] at (\xfour,\yC) {};
\node[future] at (\xfive,\yC) {};
\draw[accent,line width=1pt] (\xone-0.24,\yC-0.37) -- (\xone+0.24,\yC-0.37);

\draw[grayarr] (xc0.east) to[out=8,in=180] (fusec1.west);
\draw[grayarr] (\xone,\yC-0.37) -- (fusec1.north);
\draw[grayarr] (fusec1.south) -- (xh1c.north);

\draw[funarr]
    (xh1c.north east) to[out=28,in=200]
    node[pos=.56,above,sloped,flabel] {$f_\theta$}
    (\xtwo-0.45,\yC+0.02);

\node[used]   at (\xtwo,\yC) {};
\node[future] at (\xthree,\yC) {};
\node[future] at (\xfour,\yC) {};
\node[future] at (\xfive,\yC) {};
\draw[accent,line width=1pt] (\xtwo-0.24,\yC-0.37) -- (\xtwo+0.24,\yC-0.37);

\draw[grayarr] (xh1c.east) to[out=8,in=180] (fusec2.west);
\draw[grayarr] (\xtwo,\yC-0.37) -- (fusec2.north);
\draw[grayarr] (fusec2.south) -- (xh2c.north);

\draw[funarr]
    (xh2c.north east) to[out=28,in=200]
    node[pos=.56,above,sloped,flabel] {$f_\theta$}
    (\xthree-0.45,\yC+0.02);

\node[used]   at (\xthree,\yC) {};
\node[future] at (\xfour,\yC) {};
\node[future] at (\xfive,\yC) {};
\draw[accent,line width=1pt] (\xthree-0.24,\yC-0.37) -- (\xthree+0.24,\yC-0.37);

\draw[grayarr] (xh2c.east) to[out=8,in=180] (fusec3.west);
\draw[grayarr] (\xthree,\yC-0.37) -- (fusec3.north);
\draw[grayarr] (fusec3.south) -- (xh3c.north);

\draw[funarr]
    (xh3c.north east) to[out=28,in=200]
    node[pos=.56,above,sloped,flabel] {$f_\theta$}
    (\xfour-0.45,\yC+0.02);

\node[used]   at (\xfour,\yC) {};
\node[future] at (\xfive,\yC) {};
\draw[accent,line width=1pt] (\xfour-0.24,\yC-0.37) -- (\xfour+0.24,\yC-0.37);

\draw[grayarr] (xh3c.east) to[out=8,in=180] (fusec4.west);
\draw[grayarr] (\xfour,\yC-0.37) -- (fusec4.north);
\draw[grayarr] (fusec4.south) -- (xh4c.north);

\draw[funarr]
    (xh4c.north east) to[out=28,in=200]
    node[pos=.56,above,sloped,flabel] {$f_\theta$}
    (\xfive-0.45,\yC+0.02);

\node[used]  at (\xfive,\yC) {};
\node[faded] at (\xsix,\yC) {};
\node[expl,text=black!45] at (15.25,\yC) {$\cdots$};
\draw[accent,line width=1pt] (\xfive-0.24,\yC-0.37) -- (\xfive+0.24,\yC-0.37);

\end{tikzpicture}
\caption{
A trained $H$-output MIMO forecaster can be redeployed with smaller block size $s \le H$.
Top: default MIMO deployment uses all $H=5$ outputs in one shot.
Middle: block-recursive deployment with $s=2$ commits the first two outputs from each call; these committed predictions are fused with the current state to form the next rolled state, which is then reused by the same forecasting function $f_\theta$.
Bottom: fully recursive deployment with $s=1$ repeats the same forecast-and-roll procedure one step at a time, producing updated states $\hat{\mathbf{x}}_{t+1},\ldots,\hat{\mathbf{x}}_{t+4}$ before the final fifth committed prediction.
Points to the right of the dotted line indicate outputs emitted beyond the desired horizon; these are truncated at deployment time.
}
\label{fig:mimo_rollout_schematic}
\end{figure*}
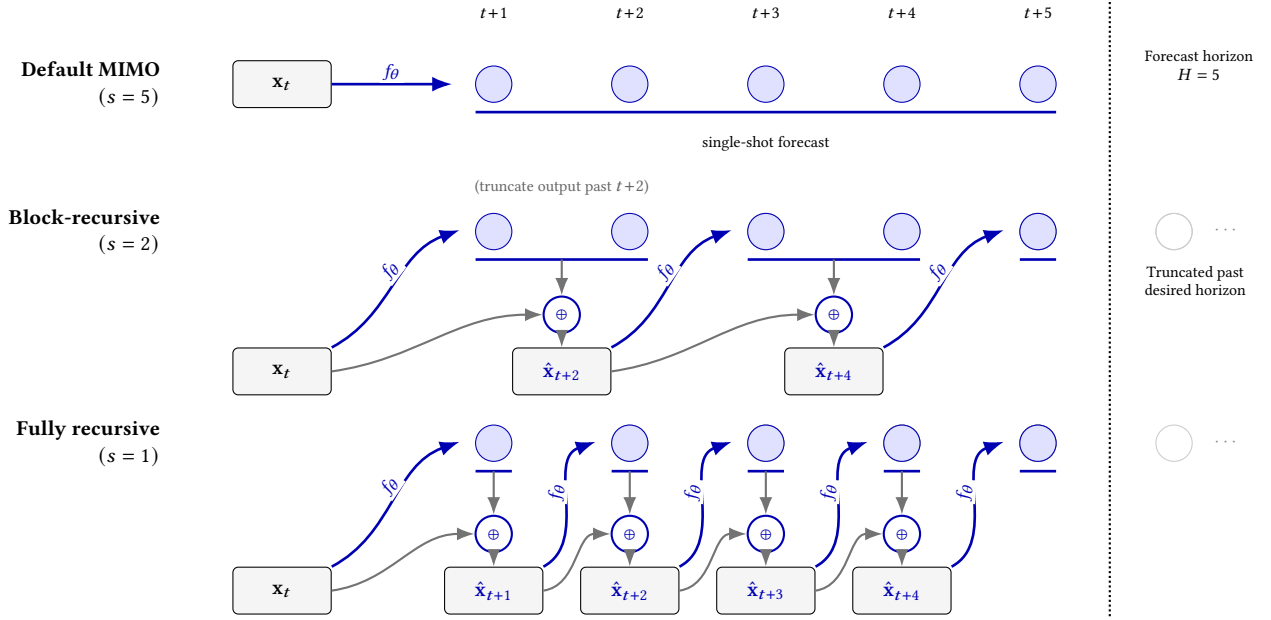

\noindent\textbf{Financial volatility forecasting.}
Volatility forecasting is central to financial risk management, volatility timing, and the broader realized-volatility literature \cite{christoffersen2000relevant,fleming2001economic,andersen2003modeling,bucci2017forecasting}. Recent surveys document the growing role of machine learning and deep learning in financial forecasting and volatility prediction \cite{wasserbacher2022machine,masini2023machine,behera2024machine,giantsidi2025deep}. Recent empirical evidence further shows that machine learning methods can be competitive with, and often outperform, strong HAR-style realized-volatility baselines \cite{christensen2023machine}. Much of this literature emphasizes model class, feature design, and predictive accuracy. Our contribution is complementary: we do not propose a new volatility architecture, but study how deployment choices within a trained multi-horizon forecaster affect accuracy, inference cost, and metric-sensitive evaluation.

\noindent\textbf{Positioning of this work.}
This paper studies deployment choice as a source of variation within a fixed trained forecaster. Rather than comparing separately trained multi-step strategies, constructing ensembles from independently trained models, or proposing a new volatility architecture, we hold the trained MIMO model fixed and vary only the inference-time rollout rule. This isolates a deployment-side mechanism by which one model induces multiple predictors with different empirical behavior. In the volatility setting, this lets us evaluate whether the induced rule family can improve MSE-oriented deployment under inference budgets, and whether policies selected under MSE remain preferable under QLIKE.

\section{Background}
\label{sec:background}

We study supervised multi-step forecasting from a univariate time series
\[
\{z_t\}_{t=1}^T.
\]
Using an input window of length $W$ and forecast horizon $H$, we form supervised input--target pairs
\begin{align}
X &= \{x_t\}_{t=W}^{T-H}, &
Y &= \{y_t\}_{t=W}^{T-H}, \\
x_t &= z_{t-W+1:t}, &
y_t &= z_{t+1:t+H}.
\end{align}
Each example therefore consists of a historical window $x_t$ and its future target block $y_t$.

A forecasting model maps the current state to an $H$-step prediction,
\[
f_\theta(x_t) \mapsto \hat y_{t+1:t+H}.
\]

A standard distinction in multi-step forecasting is between recursive and multi-output deployment. In a recursive strategy, the model predicts a small number of future steps, feeds those predictions back into the state, and repeats until the full horizon is covered. In a multi-output strategy, the model predicts the full horizon in a single call. We refer to this latter case as MIMO, which is the default direct deployment of a multi-output forecaster. Figure~\ref{fig:mimo_rollout_schematic} illustrates the relationship between default MIMO deployment, block-recursive deployment, and the fully recursive extreme; Appendix~\ref{app:worked_rollout} gives a concrete rollout example.

In our setting, a trained MIMO model can be redeployed through smaller committed output blocks. Let $s$ denote the number of forecast steps committed per call. Then
\begin{itemize}
    \item $s = H$ recovers the default MIMO forecast;
    \item $1 \leq s < H$ gives block-recursive multi-output deployment;
    \item $s = 1$ gives the fully recursive extreme.
\end{itemize}
To compare deployment rules across horizons, we index them by the rollout ratio
\[
r = \frac{s}{H}.
\]
Thus, a single trained model induces a family of deployable forecasts
\[
\left\{\hat y^{(r)}_{t+1:t+H} : r \in \mathcal R \right\}.
\]

To make this dependence explicit, let $C_s(\cdot)$ denote the operator that retains only the first $s$ outputs of an $H$-output prediction, and let $U(\cdot,\cdot)$ denote the state-update operator that appends those committed predictions to the current input window and truncates back to length $W$. Starting from $x_t^{(0)} = x_t$, deployment with block size $s$ generates rolled states
\[
x_t^{(k+1)} = U\!\left(x_t^{(k)},\, C_s\!\left(f_\theta(x_t^{(k)})\right)\right).
\]
The corresponding deployed $H$-step predictor is therefore the composed map
\[
g_{\theta,s}(x_t)
=
\operatorname{concat}\!\left(
C_s(f_\theta(x_t^{(0)})),
C_s(f_\theta(x_t^{(1)})),
\ldots
\right)_{1:H},
\]
where concatenation is truncated to the first $H$ forecast steps. Changing $s$ therefore changes the effective deployed predictor $g_{\theta,s}$ even when the trained parameters $\theta$ are fixed. This deployment-side view is consistent with \cite{green2025epistemic}, which argues that recursive composition can materially alter the effective prediction problem faced at inference time. In particular, different rollout rules evaluate the same trained forecaster on different self-induced states, and so can produce genuinely different multi-step predictors.

We study whether this induced family can be exploited through forecast aggregation. For a subset of rollout rules $S \subseteq \mathcal R$, we consider the weighted combination
\[
\hat y^{\mathrm{ens}} = \sum_{r \in S} w_r \hat y^{(r)}.
\]
The two ensemble variants differ only in how the weights are chosen. In the simple ensemble, the weights are fixed uniformly,
\[
w_r = |S|^{-1}, \qquad r \in S.
\]
In the linear ensemble, the weights $\{w_r\}_{r \in S}$ are learned from held-out input--target pairs $(x_t, y_t)$, allowing the contribution of each induced forecast to adapt to empirical performance.

The rationale for aggregation is forecast diversity, but diversity alone is not sufficient. Aggregation helps when the component forecasts make sufficiently different errors, so that averaging or weighting can cancel residual variation. When errors are strongly aligned, however, combining more forecasts may offer little benefit \cite{wood2023unified}. This is why more forecasts are not necessarily better, and why subset selection or learned weighting may outperform naive expansion of the ensemble.

\section{Experimental Setup}
\label{sec:setup}

We evaluate whether alternative deployment rules from a single trained multi-output forecaster can improve daily stock-volatility forecasting, and whether aggregating these induced forecasts yields a useful accuracy--cost trade-off.

\subsection{Data and task}
We study multi-step forecasting of daily stock volatility using the VOLARE archive \cite{cipollini2026volatility} on 20 univariate stock series, each evaluated with 3 independent repeats. The volatility target is the log realized variance, computed as a subsampled 5-minute realized variance estimator ($\mathrm{rv5\_ss}$), which aggregates squared intraday returns and provides a noise-robust proxy of daily volatility.

Each example uses an input window of length $W = 20$ to predict the next $H$ steps, with $H \in \{10, 20, 30\}$, allowing us to evaluate robustness across different forecasting horizons.

Series are 2,827 values in total length and split chronologically into train, validation, and test segments using 70\%, 15\%, 15\% for each split respectively. All subset selection and linear-combiner fitting use validation only; test is used only for final evaluation.

\begin{figure*}[t]
    \centering

    \begin{subfigure}[t]{0.33\textwidth}
        \centering
        \includegraphics[width=\linewidth]{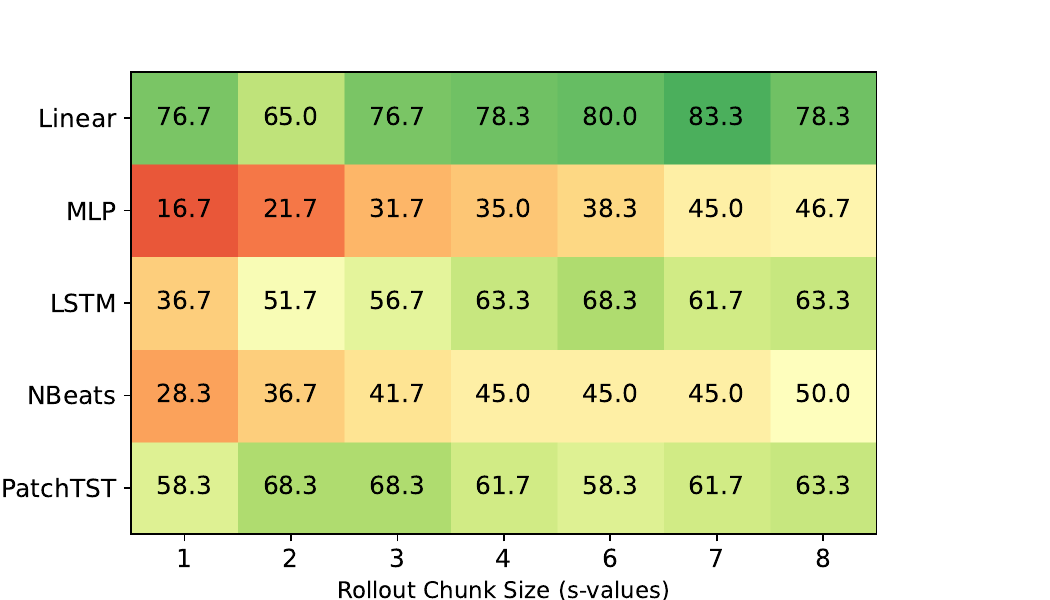}
        \caption{H = 10.}
    \end{subfigure}
    \hfill
    \begin{subfigure}[t]{0.33\textwidth}
        \centering
        \includegraphics[width=\linewidth]{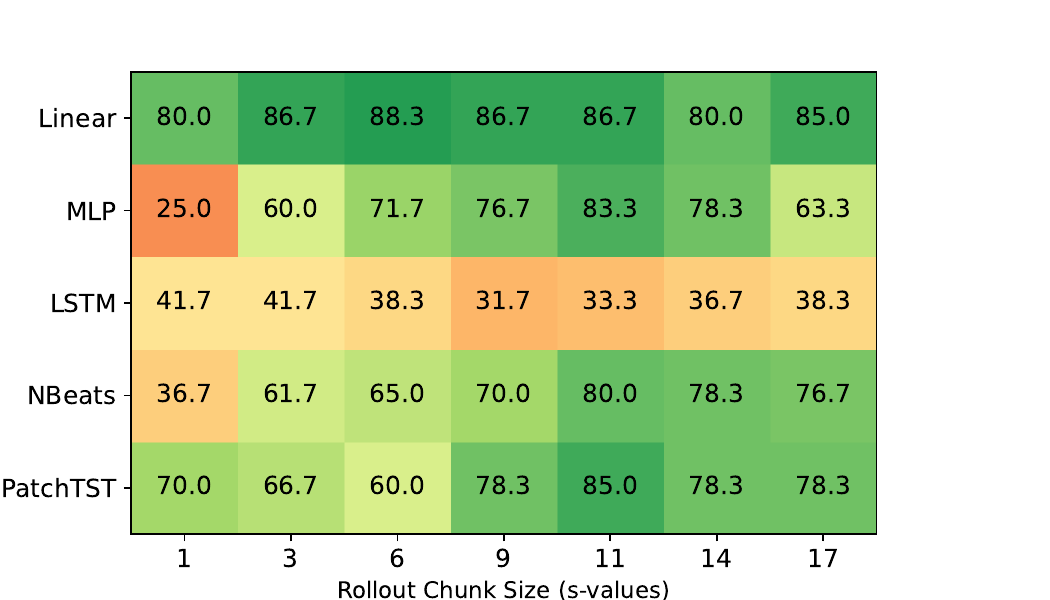}
        \caption{H = 20.}
    \end{subfigure}
    \hfill
    \begin{subfigure}[t]{0.33\textwidth}
        \centering
        \includegraphics[width=\linewidth]{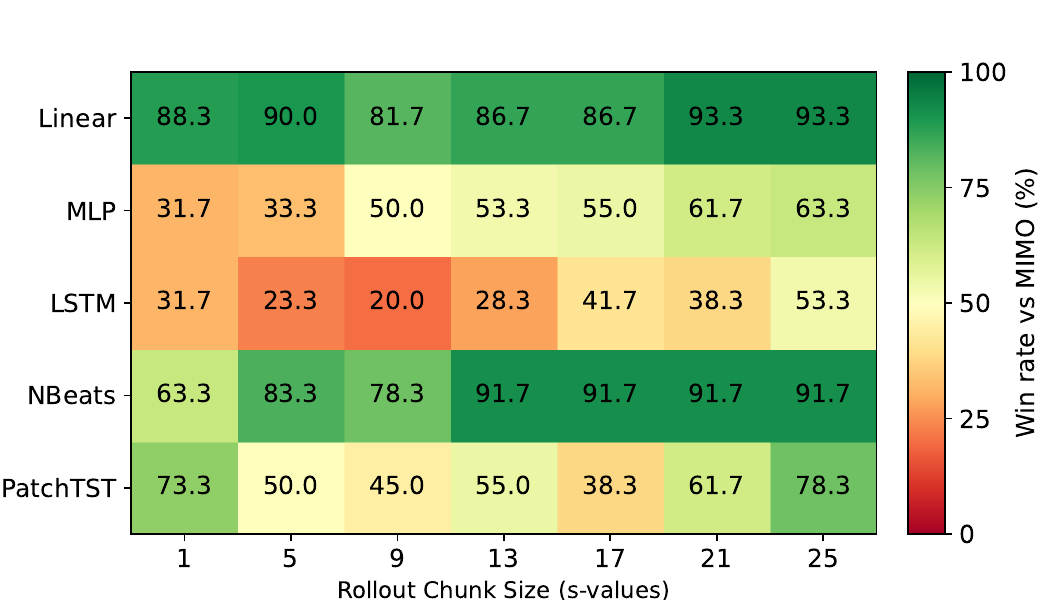}
        \caption{H = 30.}
    \end{subfigure}

    \caption{Rule-level win rates against default MIMO across horizons. For each horizon, cells show the fraction of tasks on which a non-MIMO singleton deployment rule outperforms default MIMO for the same trained model. Many non-MIMO rules beat MIMO on a majority of tasks, showing substantial deployment headroom. At the same time, the strongest rule depends on model class and horizon, indicating that no single fixed alternative rule dominates globally.}
    \label{fig:non_mimo_pcts}
\end{figure*}

\begin{figure*}[t]
    \centering

    \begin{subfigure}[t]{0.33\textwidth}
        \centering
        \includegraphics[width=\linewidth]{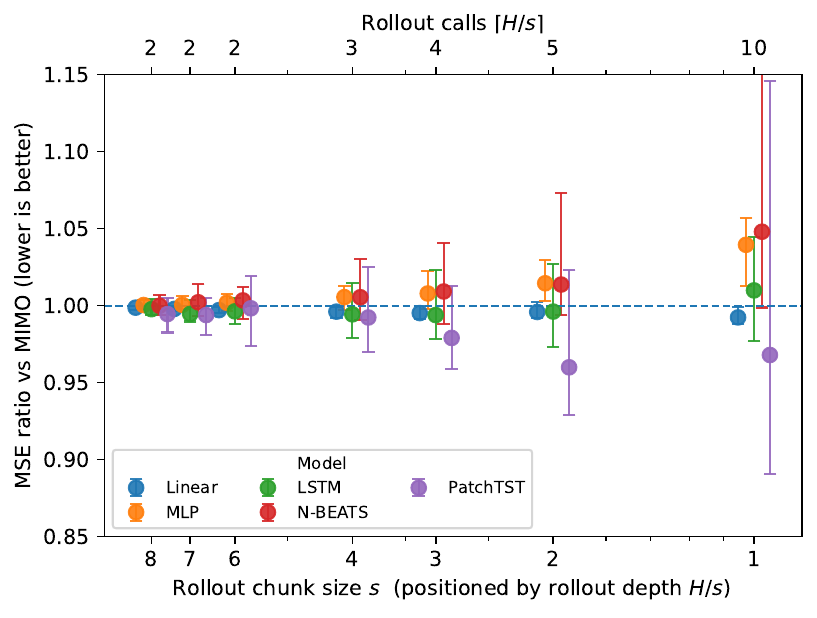}
        \caption{H = 10.}
    \end{subfigure}
    \hfill
    \begin{subfigure}[t]{0.33\textwidth}
        \centering
        \includegraphics[width=\linewidth]{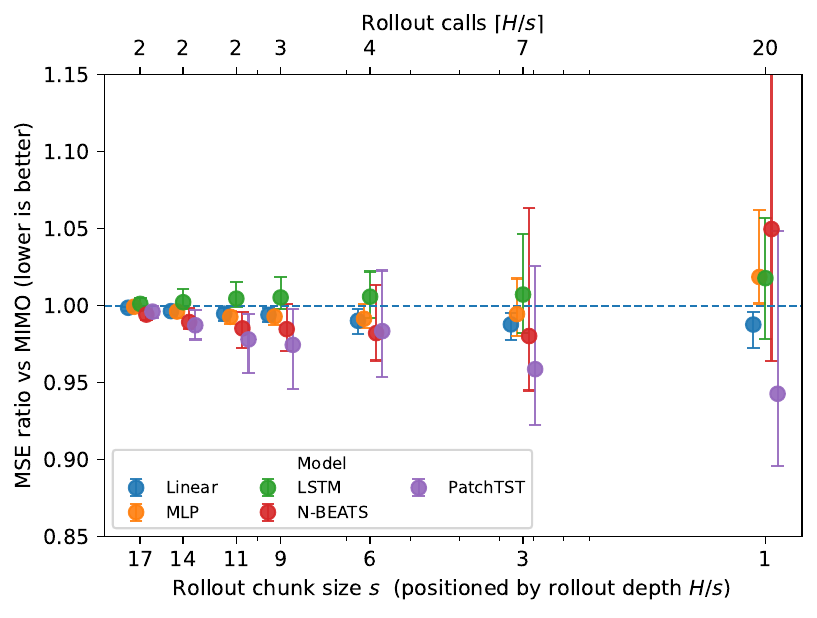}
        \caption{H = 20.}
    \end{subfigure}
    \hfill
    \begin{subfigure}[t]{0.33\textwidth}
        \centering
        \includegraphics[width=\linewidth]{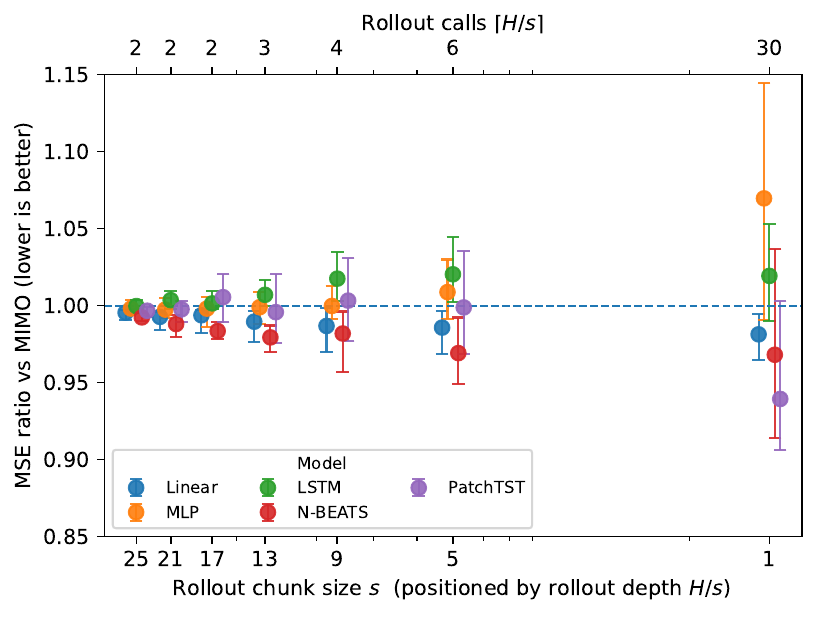}
        \caption{H = 30.}
    \end{subfigure}

\caption{\textbf{Singleton deployment rules expose a heterogeneous accuracy--cost landscape.}
Each point is a fixed non-MIMO singleton rule for a given model class. The x-position uses the pre-ceiling rollout depth $H/s$ to separate rules with different chunk sizes, while the top axis reports the actual call budget $\lceil H/s\rceil$. The y-axis shows the median test MSE ratio relative to default MIMO, with error bars denoting interquartile variation across tasks. Dashed lines mark MIMO parity in error.}
\label{fig:fixed_rule_tradofff}
\end{figure*}

\subsection{Models}

We evaluate a diverse set of forecasting architectures spanning linear, recurrent, and transformer-based models. Specifically, we consider a linear baseline (\textbf{LinearNet}), a multilayer perceptron (\textbf{MLP}), a long short-term memory recurrent network (\textbf{LSTM} \cite{hochreiter1997long}), a deep residual forecasting architecture (\textbf{N-BEATS} \cite{oreshkin2019n}), and a transformer-based time series model (\textbf{PatchTST} \cite{nie2022time}).

Each model is trained as a direct multi-output forecaster, predicting the full horizon in a single forward pass. Training is performed using the Adam optimizer with learning rate $10^{-3}$, batch size 128, and 100 epochs. Model selection is based on validation error over the full forecasting horizon. Additional architecture-specific hyperparameters are provided in Appendix~\ref{app:further_exp_details}.

\subsection{Induced rule pool and deployable policies}
For each trained model, we evaluate a fixed pool of inference-time rollout rules. Each rule is defined by an output block size $s$, or equivalently a rollout ratio $r=s/H$, and produces a distinct multi-step forecast from the same trained parameters. For each horizon $H \in \{10,20,30\}$, we use eight approximately uniformly spaced output sizes between $1$ and $H$. This gives coverage across a broad range of recursive depth and inference-time cost, from the fully recursive extreme ($s=1$) to the default direct MIMO forecast ($s=H$).

We evaluate the following deployment policies:
\begin{itemize}
    \item \textbf{Default MIMO}: the standard direct forecast with $s=H$;
    \item \textbf{Singleton rules}: each induced rule deployed on its own;
    \item \textbf{Validation-selected singleton}: the single induced rule with the best validation performance;
    \item \textbf{Subset ensembles}: simple or linear ensembles formed from subsets of induced rules;
    \item \textbf{Full simple ensemble}: the uniform average over the full rule pool;
    \item \textbf{Full linear ensemble}: a linear combination over the full rule pool.
\end{itemize}

To study the budgeted deployment space, we evaluate ensembles built from rule subsets of size \(k\). 
The reported \(k\)-subset policies use greedy forward selection on the validation split. Starting from the empty set, we add one rollout rule at a time. At each step, we choose the remaining rule whose inclusion gives the lowest validation MSE for the current ensemble. For a greedy simple subset, the validation score is the MSE of the uniform average over the selected rules. For a greedy linear subset, the validation score is the MSE after fitting an ordinary-least-squares linear combiner on the validation split, with no intercept. After \(k\) rules have been selected, the selected subset is fixed and evaluated once on the test split. Linear subset weights are fitted only on validation data and then applied unchanged to the test forecasts.

We also evaluate randomly sampled \(k\)-subsets as diagnostic baselines. For each \(k\), we sample up to
$
\min\left\{\binom{|\mathcal{R}_{\mathrm{cand}}|}{k}, 50\right\}$
distinct subsets uniformly without replacement from the candidate rule pool \(\mathcal{R}_{\mathrm{cand}}\), enumerating all subsets when fewer than 50 are available.

\subsection{Evaluation}

Our primary baseline is the default MIMO deployment within each function class. Since the central question is whether a trained multi-output forecaster is best used through its default MIMO rule, we report all main metrics relative to this within-class baseline.

Let $y_{t+h} = \log \mathrm{RV}_{t+h}$ denote the target at horizon step $h$, where $\mathrm{RV}_{t+h}$ is the realized variance, and let $\hat y_{t+h}$ denote the corresponding forecast. Performance is evaluated at the task level over the full multi-step horizon, rather than stepwise by forecast depth, so that each horizon forecast is treated as a single decision problem.

For a single forecast block, we define the horizon-averaged mean squared error on the log-volatility scale as
\[
\mathrm{MSE}(\hat y_t, y_t)
=
\frac{1}{H}\sum_{h=1}^{H} (\hat y_{t+h} - y_{t+h})^2.
\]

Because the forecasting target is log realized variance, QLIKE is evaluated on the original variance scale rather than directly on the log scale. Defining
\[
v_{t+h} = \exp(y_{t+h}), \qquad \hat v_{t+h} = \exp(\hat y_{t+h}),
\]
we compute the horizon-averaged QLIKE loss as
\[
\mathrm{QLIKE}(\hat y_t, y_t)
=
\frac{1}{H}\sum_{h=1}^{H}
\left(
\frac{v_{t+h}}{\hat v_{t+h}}
-
\log \frac{v_{t+h}}{\hat v_{t+h}}
-
1
\right).
\]
Equivalently, since $y_{t+h}=\log v_{t+h}$ and $\hat y_{t+h}=\log \hat v_{t+h}$,
\[
\mathrm{QLIKE}(\hat y_t, y_t)
=
\frac{1}{H}\sum_{h=1}^{H}
\left(
\exp(y_{t+h}-\hat y_{t+h})
-
(y_{t+h}-\hat y_{t+h})
-
1
\right).
\]

For a deployed predictor, each loss is averaged over all test forecast blocks in the task. Let $L(\hat y)$ denote the resulting value of an evaluation metric for a deployed forecast $\hat y$, and let $\hat y^{\mathrm{MIMO}}$ denote the corresponding default MIMO forecast. For any loss or cost metric $L$, we report the relative score
\[
\frac{L(\hat y)}{L(\hat y^{\mathrm{MIMO}})}.
\]
Values below $1$ indicate improvement over default MIMO, while values equal to $1$ indicate parity.

Using this convention, we report relative test MSE, relative QLIKE, and relative inference time. MSE evaluates predictive accuracy on the log-realized-variance scale used for training, while QLIKE evaluates the forecasts on the original variance scale and is standard in volatility forecasting. Win rate versus MIMO denotes the fraction of tasks on which a deployment policy achieves lower test MSE than default MIMO. For ensemble policies, inference time is taken as the sum of the inference costs of the constituent induced forecasts; we ignore aggregation overhead and linear weight fitting cost, since our focus is deployment-time forecasting cost and these additional costs are negligible in comparison.



\section{Results}
\label{sec:results}

The empirical results are organized around the deployment decisions a practitioner would face after training a MIMO forecaster. We first ask whether the induced rollout rules contain useful alternatives to default MIMO deployment. We then ask whether validation-based policies can exploit those alternatives under inference budgets. Finally, we examine two qualifications: whether policies selected under MSE remain preferable under QLIKE, and whether induced policies are competitive with policies built from separately trained rule-specific models.

Figures~\ref{fig:non_mimo_pcts} and~\ref{fig:fixed_rule_tradofff} summarize the rule-level evidence. They show that the induced rule family contains useful deployment headroom, but also that this headroom is heterogeneous across architectures, horizons, and inference costs. The remaining results therefore focus on validation-based policies rather than on choosing a single fixed non-MIMO rule.

\begin{figure}[b]
    \centering
    \includegraphics[width=\linewidth]{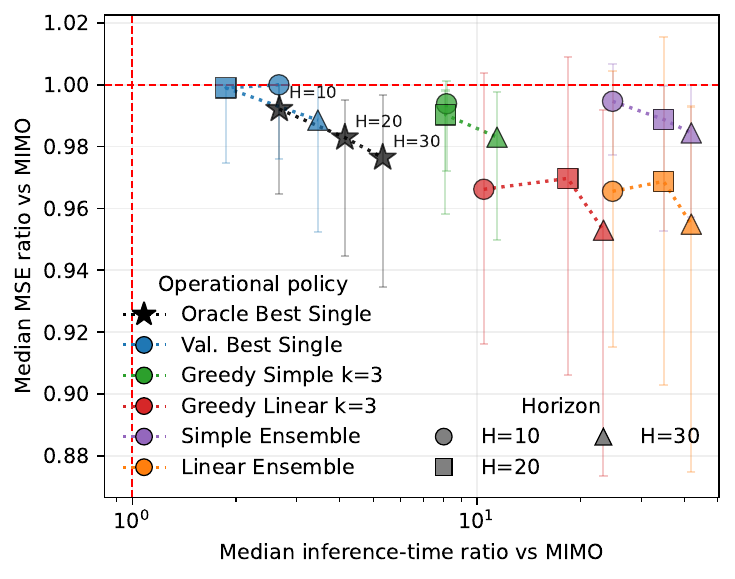}
    \caption{\textbf{Operational deployment policies improve the MSE--cost trade-off over default MIMO.}
Each point summarizes a policy by its median inference-time ratio and median test MSE ratio relative to MIMO, pooled over model classes; marker shape indicates horizon. Dashed lines mark the unit-cost, unit-error MIMO baseline, and black stars show the oracle best singleton benchmark. Validation-selected singletons offer low-cost gains, while subset and ensemble policies achieve larger MSE reductions at higher inference cost, with linear aggregation often matching or exceeding the oracle singleton reference.}
    \label{fig:operational_mse_vs_time_h}
\end{figure}

\begin{table*}[t]
\centering
\small
\setlength{\tabcolsep}{3.5pt}
\resizebox{\textwidth}{!}{%
\begin{tabular}{lcccccccccccccccccccc}
\toprule
& \multicolumn{4}{c}{Val-sin} 
& \multicolumn{4}{c}{Sim-k3} 
& \multicolumn{4}{c}{Lin-k3} 
& \multicolumn{4}{c}{Sim-ens} 
& \multicolumn{4}{c}{Lin-Ens} \\
\cmidrule(lr){2-5}
\cmidrule(lr){6-9}
\cmidrule(lr){10-13}
\cmidrule(lr){14-17}
\cmidrule(lr){18-21}
Model 
& Win & Median & LQ & UQ 
& Win & Median & LQ & UQ 
& Win & Median & LQ & UQ 
& Win & Median & LQ & UQ 
& Win & Median & LQ & UQ \\
\midrule
Linear   
& 65.6 & 0.995 & 0.979 & 1.000
& 93.3 & 0.994 & 0.984 & 0.998
& 70.0 & 0.973 & 0.948 & 1.005
& 92.2 & 0.993 & 0.982 & 0.997
& 68.3 & 0.974 & 0.949 & 1.004 \\

MLP      
& 27.8 & 1.000 & 0.999 & 1.000
& 65.6 & 0.998 & 0.989 & 1.002
& 65.0 & 0.981 & 0.957 & 1.025
& 57.8 & 0.995 & 0.984 & 1.009
& 64.4 & 0.983 & 0.958 & 1.025 \\

LSTM     
& 35.6 & 1.000 & 0.990 & 1.020
& 48.9 & 1.000 & 0.987 & 1.015
& 51.1 & 0.999 & 0.973 & 1.038
& 49.4 & 1.000 & 0.987 & 1.015
& 50.6 & 0.999 & 0.974 & 1.038 \\

NBEATS   
& 57.8 & 0.991 & 0.968 & 1.000
& 85.6 & 0.975 & 0.947 & 0.993
& 85.0 & 0.942 & 0.908 & 0.980
& 73.3 & 0.970 & 0.947 & 1.003
& 85.0 & 0.943 & 0.907 & 0.980 \\

PatchTST 
& 86.7 & 0.945 & 0.901 & 0.989
& 95.6 & 0.950 & 0.920 & 0.971
& 95.6 & 0.838 & 0.808 & 0.874
& 81.7 & 0.966 & 0.946 & 0.986
& 95.6 & 0.837 & 0.808 & 0.876 \\
\bottomrule
\end{tabular}%
}
\caption{\textbf{Test performance of deployable policies by function class, relative to default MIMO.}
For each model class and policy, we report the fraction of tasks on which the policy improves over MIMO (Win, \%), together with the median, lower quartile (LQ), and upper quartile (UQ) of the test MSE ratio relative to MIMO. Values below 1 indicate improvement over default MIMO. Aggregation often improves median MSE, but the strongest deployable policy remains architecture-dependent. }
\label{tab:fused_results}

\end{table*}

\subsection{Inference-time rollout rules expose deployment headroom}

Figure~\ref{fig:non_mimo_pcts} shows that non-MIMO singleton rules frequently outperform default MIMO. The effect is not uniform: the strongest chunk size varies across model classes and horizons, and some architectures admit more useful non-MIMO alternatives than others. Thus, default MIMO is often not the best deployment of a trained multi-output model, but no single fixed rollout rule is a reliable global replacement.

Figure~\ref{fig:fixed_rule_tradofff} adds the cost dimension. Singleton rules occupy different regions of the accuracy--cost landscape: some deeper rollouts improve over MIMO, others do not, and rules with similar rollout depth can have different median MSE ratios. This indicates that the induced rule family contains useful deployment headroom, but inference cost alone is not a sufficient selection criterion.




\subsection{Validation-based policies exploit the induced rule family}

We next ask whether validation-based policies can use this heterogeneous rule family effectively. Figure~\ref{fig:operational_mse_vs_time_h} shows that, pooled over model classes, most operational policies improve on the MIMO baseline in median test MSE. Validation-selected singletons give low-cost gains, while subset and ensemble policies typically achieve larger reductions at higher inference cost. Linear subset and full linear aggregation are among the strongest pooled policies, often matching or exceeding the oracle best singleton benchmark.

Table~\ref{tab:fused_results} shows that these gains remain architecture-dependent. Linear models benefit most consistently from simple averaging policies, while MLP and LSTM are closer to parity. N-BEATS and PatchTST show larger gains from aggregation, especially learned linear aggregation. Overall, validation-based deployment policies improve the MSE--cost trade-off, but the most effective policy depends on the model class.

\subsection{The preferred operational policy depends on the budget}

Table~\ref{tab:policy_vs_mimo_grouped} summarizes the pooled MSE--cost trade-off by horizon. Validation-selected singletons are the lowest-cost upgrade over MIMO, but their accuracy gains are modest. Small subsets provide a stronger middle ground: simple $k=3$ subsets improve win rates at moderate cost, while linear $k=3$ subsets achieve the largest median MSE reductions in the pooled results. Full ensembles are substantially more expensive, and their additional gains over the corresponding $k=3$ policies are often small.

This pattern is most visible for linear aggregation. At $H=10$ and $H=20$, the full linear ensemble and linear $k=3$ subset have nearly identical median MSE ratios, but the subset is much cheaper. At $H=30$, the full linear ensemble again provides only a marginal additional gain at considerably higher cost. 
These results suggest a simple operational protocol. A practitioner can first evaluate the induced rule pool on validation data, then choose the cheapest policy that meets the required accuracy improvement over default MIMO. When additional budget is minimal, this favors a validation-selected singleton; when moderate extra inference is acceptable, small subsets offer a stronger accuracy--cost compromise; and when cost is secondary, full ensembles serve as high-cost endpoints.

\begin{table}[t]
\small
\centering
\begin{tabular}{lcccc}
\toprule
Policy & Win rate (\%) & MSE ratio & Time ratio \\
\midrule
\multicolumn{4}{c}{Horizon = 10} \\
\midrule
Val-sin & 44.3 & 1.000 [0.976, 1.000] & 2.66 [1.00, 5.46] \\
Sim-k3  & 70.7 & 0.994 [0.972, 1.001] & 8.16 [4.80, 12.18] \\
Lin-k3  & 73.0 & 0.966 [0.916, 1.004] & 10.47 [8.11, 13.22] \\
Sim-ens & 62.7 & 0.995 [0.977, 1.007] & 24.76 [20.14, 25.72] \\
Lin-ens & 72.0 & 0.966 [0.915, 1.004] & 24.76 [20.14, 25.72] \\
\midrule
\multicolumn{4}{c}{Horizon = 20} \\
\midrule
Val-sin & 51.3 & 0.999 [0.975, 1.000] & 1.87 [1.00, 6.68] \\
Sim-k3  & 82.3 & 0.990 [0.958, 0.998] & 8.10 [4.73, 19.64] \\
Lin-k3  & 70.0 & 0.970 [0.906, 1.009] & 18.35 [13.44, 22.28] \\
Sim-ens & 75.0 & 0.989 [0.953, 0.999] & 34.83 [27.74, 36.26] \\
Lin-ens & 69.0 & 0.969 [0.903, 1.015] & 34.83 [27.74, 36.26] \\
\midrule
\multicolumn{4}{c}{Horizon = 30} \\
\midrule
Val-sin & 68.3 & 0.989 [0.952, 1.000] & 3.45 [1.64, 18.92] \\
Sim-k3  & 80.3 & 0.983 [0.950, 0.998] & 11.44 [5.73, 29.97] \\
Lin-k3  & 77.0 & 0.953 [0.874, 0.992] & 23.30 [11.60, 31.13] \\
Sim-ens & 75.0 & 0.984 [0.957, 1.000] & 41.91 [32.95, 43.84] \\
Lin-ens & 77.3 & 0.955 [0.875, 0.993] & 41.91 [32.95, 43.84] \\
\bottomrule
\end{tabular}
\caption{\textbf{Pooled policy-level performance relative to default MIMO.}
For each horizon and policy, we report the win rate against MIMO, the median test MSE ratio, and the median inference-time ratio, with interquartile ranges in brackets. Results are aggregated over model classes. Values below 1 indicate improved MSE or lower cost relative to MIMO. The table shows a deployment ladder: validation-selected singletons are the lowest-cost upgrade, small subsets provide larger gains at moderate cost, and full ensembles are the highest-cost endpoint.}
\label{tab:policy_vs_mimo_grouped}
\end{table}

\subsection{MSE gains do not transfer equally to QLIKE across deployment policies}

The preceding results use test MSE relative to default MIMO as the primary objective. We now evaluate whether the same policies also improve QLIKE. To isolate transfer rather than re-selection, all policies are selected exactly as before using validation MSE only, and are then evaluated on the test set under both MSE and QLIKE.

\begin{figure}[t]
    \centering
    \includegraphics[width=\linewidth]{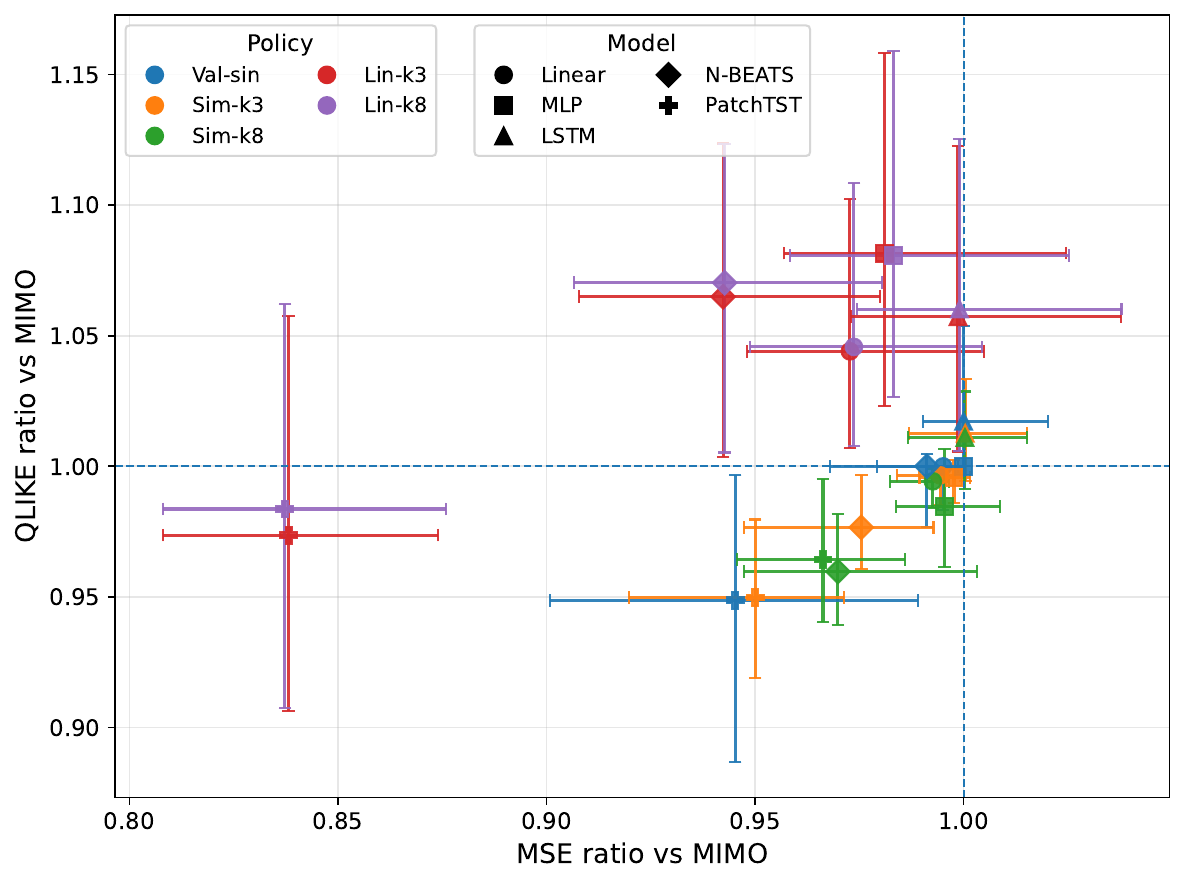}
    \caption{
\textbf{Transfer of MSE-selected deployment policies to QLIKE.}
Each point is a model-class--policy pair, with median test MSE ratio on the x-axis and median test QLIKE ratio on the y-axis, both relative to default MIMO. Error bars show interquartile ranges across tasks. Policies are selected using validation MSE only and are not re-tuned for QLIKE. Dashed lines mark MIMO parity; lower-left points improve both metrics, while upper-left points improve MSE but worsen QLIKE. 
}
    \label{fig:policy_transfer_qlike}
\end{figure}

Figure~\ref{fig:policy_transfer_qlike} shows the model-class--policy level comparison. Several policies improve both MSE and QLIKE, but transfer is heterogeneous. Validation-selected singletons and simple averaging policies more often remain near or inside the joint-improvement region. Learned linear combinations achieve stronger MSE reductions in several cases, but their QLIKE improvements are less consistent.

\begin{figure}[hbt]
    \centering
    \includegraphics[width=\linewidth]{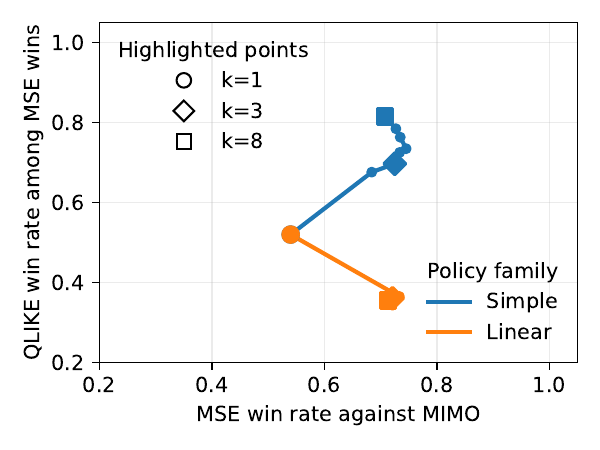}
    \caption{
\textbf{Transfer of MSE wins to QLIKE by deployment family.}
Each point corresponds to a subset size $k$, aggregated across architectures within a deployment family. The x-axis shows the MSE win rate against default MIMO, and the y-axis shows the conditional QLIKE win rate among those MSE wins. Simple and linear subsets have similar MSE win rates, but simple subsets transfer MSE wins into QLIKE wins.
}
    \label{fig:transfer_rate_qlike}
\end{figure}

Figure~\ref{fig:transfer_rate_qlike} summarizes the same pattern by deployment family. Simple and linear subsets have broadly similar MSE win rates, but simple subsets convert MSE wins into QLIKE wins more reliably. Thus, policies that are strongest under squared error are not necessarily the most reliable under QLIKE. 
The point is not that MSE-based deployment is invalid. Rather, MSE and QLIKE reward different aspects of volatility forecasts, so a policy selected for one objective may not be the preferred policy for another. This makes deployment selection part of the evaluation design: the induced rule family should be searched under the operational loss that will matter in the intended financial use case.

\subsection{Induced policies avoid a separately trained rule sweep}

Finally, we compare induced policies with a more expensive alternative: training a separate model for each candidate rollout block size and applying the same validation-based policy class to that trained rule family. This comparison does not assume rule-wise equivalence. An induced rule with block size $s$ and a model trained specifically for block size $s$ are different predictors, and the best rule in each family need not coincide. The question is instead practical: how much performance is lost by using the rule family exposed by one trained MIMO model, rather than paying for a separately trained rule sweep?

Table~\ref{tab:induced_vs_trained_policy_cost} reports
$\Delta_{\mathrm{MSE}} =
\frac{\mathrm{MSE}_{\mathrm{induced}} - \mathrm{MSE}_{\mathrm{trained}}}
{\mathrm{MSE}_{\mathrm{trained}}}$,
so negative values favor induced policies and zero denotes parity. Separately trained rule families are slightly stronger overall, as reflected by induced win rates below $50\%$ for most policies. However, median $\Delta_{\mathrm{MSE}}$ remains close to zero, so the extra training cost yields little median MSE improvement. Even under the oracle post-selection cost reported in the table, the induced-to-trained training ratio is about $0.30$ for greedy $k=3$ policies and $0.11$ for full ensembles. Under full policy search, the contrast is larger: induced policies require one trained MIMO model, whereas the separately trained alternative requires one model per candidate block size.
\begin{table}[t]
\small
\centering
\setlength{\tabcolsep}{4pt}
\begin{tabular}{lccccc}
\toprule
\textbf{Policy} & \textbf{Induced Win (\%)} & \textbf{Median $\Delta_{\mathrm{MSE}}$} & \textbf{Q25} & \textbf{Q75} & \textbf{\makecell{Oracle \\ Ratio}} \\
\midrule
Val-sin   & 50.2 & 0.00 & -0.02 & 0.02 & 0.93 \\
G-Sim-3  & 28.2 & 0.01 & 0.00 & 0.04 & 0.30 \\
G-Lin-3  & 40.7 & 0.00 & -0.01 & 0.02 & 0.30 \\
Sim-ens  & 23.8 & 0.01 & 0.00 & 0.04 & 0.11 \\
Lin-ens  & 41.8 & 0.00 & -0.01 & 0.02 & 0.11 \\
\bottomrule
\end{tabular}
\caption{
\textbf{Induced policies versus separately trained rule families.}
We report $\Delta_{\mathrm{MSE}} =
(\mathrm{MSE}_{\mathrm{induced}}-\mathrm{MSE}_{\mathrm{trained}})/
\mathrm{MSE}_{\mathrm{trained}}$, where negative values favor induced policies. ``Induced Win'' is the fraction of runs where the induced policy has lower MSE. ``Oracle Ratio'' is a post-selection lower bound on $\mathrm{Train}_{\mathrm{induced}}/\mathrm{Train}_{\mathrm{trained}}$, charging the trained alternative only for the final selected policy members rather than for the full rule sweep.
}
\label{tab:induced_vs_trained_policy_cost}
\end{table}

\section{Discussion}
\label{sec:discussion}

Our results show that deployment is an active part of predictive design in multi-horizon financial forecasting. A trained multi-output forecaster does not define only one deployed system: alternative rollout rules induce different predictors from the same learned parameters, with different empirical accuracy and cost profiles. This distinction matters in volatility forecasting because predictions are generated repeatedly across assets and horizons, often under inference budgets and under evaluation criteria that reflect downstream financial use.

The main practical implication is that default MIMO deployment is often not the best use of a trained MIMO model. Non-default rollout rules frequently improve over MIMO, but the strongest fixed rule depends on the model class and forecast horizon. This makes a single global replacement rule unreliable and instead motivates deployment policies over the induced rule family. Under the primary MSE objective, validation-selected singletons provide a low-cost upgrade, while small subsets recover much of the available gain at substantially lower cost than full ensembles.

The QLIKE results qualify this MSE-based conclusion rather than replacing it. Policies selected by validation MSE do not transfer uniformly to QLIKE: learned linear combinations often achieve the strongest MSE reductions, but their improvements are less consistent under the volatility-sensitive loss. Simpler policies, such as singleton selection and simple averaging, tend to transfer more reliably across the two metrics. Thus, the preferred deployment policy depends not only on inference budget, but also on the operational loss used to evaluate volatility forecasts.

Overall, the results support viewing a trained MIMO forecaster as a compact deployment space rather than a single predictor. This space can be exploited with simple validation-based policies, but the best operating point is budget- and metric-dependent. In financial forecasting settings where repeated train and inference cost, and evaluation mismatch are both relevant, deployment policy should therefore be evaluated alongside model architecture. Practically, the results suggest a lightweight deployment check for multi-horizon volatility forecasters. After training a MIMO model, one can generate a small pool of induced rollout policies, evaluate them on validation data under the relevant operational loss, and select the cheapest policy that achieves the desired improvement over default MIMO. This procedure does not require retraining the base model, but it can reveal whether the trained forecaster is being underused by its default deployment rule.

\subsubsection*{Limitations and Future Work}
\label{sec:limitations}

Our study is limited to univariate daily realized-volatility forecasting, a fixed rollout family based on output block size, and simple deployment policies based on selection and aggregation. The quantitative trade-offs may differ in multivariate, higher-frequency, or cross-asset settings, and stronger budget-aware policies may improve on the baselines considered here. 
We evaluate forecast losses rather than downstream decisions; our results should not be read as evidence of improved portfolio utility or risk-control performance.
They show that deployment policies can change volatility-forecast accuracy, cost, and QLIKE behavior; whether these changes translate into better financial decisions remains an important downstream evaluation question.

\subsection*{Conclusion}
\label{sec:conclusion}

A trained MIMO volatility forecaster is not a single deployed predictor, but a family of predictors induced by inference-time rollout choices. We showed that this deployment space often improves over default MIMO, with validation-selected singletons and small subsets offering practical MSE--cost trade-offs without a separately trained rule sweep. Financial forecasting systems should therefore evaluate not only what model is trained, but how that model is deployed.

\bibliographystyle{ACM-Reference-Format}
\bibliography{concise_references}

\FloatBarrier
\appendix

\section{Additional Methodological Context}
\label{app:method_context}

This appendix adds a few extra notes on the forecasting target, the evaluation losses, the baseline comparison, and the rollout construction. These details are included to make the experimental setup easier to interpret.

\subsection{Log realized variance and QLIKE}
\label{app:qlike_intuition}

We model log realized variance rather than realized variance directly because realized variance is strictly positive, highly right-skewed, and typically more stable on the logarithmic scale. The log transformation reduces the effect of extreme volatility days and makes squared-error training closer to a scale-stabilized regression problem.

QLIKE is evaluated on the original variance scale and is commonly used in volatility forecasting because it penalizes errors in variance forecasts in a way that is relevant for risk measurement. Writing $v$ for realized variance and $\hat v$ for the variance forecast, the QLIKE loss is
\[
\frac{v}{\hat v} - \log\!\left(\frac{v}{\hat v}\right) - 1.
\]
This loss is minimized at $\hat v=v$, but its shape is asymmetric in economically meaningful ways. In particular, severe under-prediction of variance can be heavily penalized because the ratio $v/\hat v$ becomes large. This differs from log-scale MSE, which treats positive and negative log-errors symmetrically. Consequently, a deployment policy selected to minimize validation MSE need not be the best policy under QLIKE, which motivates the metric-transfer analysis in Section~\ref{sec:results}.

\subsection{Scope of the baseline comparison}
\label{app:har_scope}

The central comparison in this paper is between deployment policies for a fixed trained multi-output forecaster, not between volatility model classes. We therefore use within-class default MIMO as the primary baseline. This isolates the deployment-side question: once the model and trained parameters are fixed, is the standard direct MIMO forecast the best operational use of that model?

Classical realized-volatility models such as HAR-RV are important baselines for absolute forecasting performance, but they address a different question: whether a machine-learning architecture outperforms a classical volatility model. HAR-style benchmarking would help calibrate absolute forecast quality, but it would not directly test whether alternative rollout rules improve a fixed MIMO forecaster. We therefore treat it as complementary to the within-class deployment comparisons studied here.

\subsection{Worked example of induced rollout}
\label{app:worked_rollout}

Consider a trained MIMO forecaster with horizon $H=10$ and a deployment block size $s=2$. On the first call, the model produces a 10-step forecast
\[
f_\theta(x_t) =
(\hat y_{t+1}^{(0)}, \ldots, \hat y_{t+10}^{(0)}).
\]
The deployment rule commits only the first two values,
\[
(\hat y_{t+1}^{(0)}, \hat y_{t+2}^{(0)}),
\]
appends them to the input window, and truncates the window back to length $W$ to obtain the rolled state $x_t^{(1)}$. The same trained model is then evaluated again:
\[
f_\theta(x_t^{(1)}) =
(\hat y_{t+3}^{(1)}, \ldots, \hat y_{t+12}^{(1)}),
\]
where only the first two outputs are again committed. Repeating this process five times produces the deployed 10-step forecast
\[
\hat y_{t+1:t+10}^{(s=2)}
=
(\hat y_{t+1}^{(0)},\hat y_{t+2}^{(0)},
\hat y_{t+3}^{(1)},\hat y_{t+4}^{(1)},
\ldots,
\hat y_{t+9}^{(4)},\hat y_{t+10}^{(4)}).
\]
The trained parameters are unchanged throughout. Only the inference-time state sequence changes. Thus, different choices of $s$ define different deployed predictors even though they reuse the same trained MIMO forecasting map.

\section{Additional experimental details}
\label{app:further_exp_details}

\begin{table*}[t]
\centering
\small
\setlength{\tabcolsep}{2.1pt}
\renewcommand{\arraystretch}{0.95}
\begin{tabular}{llrrrrrrrrrrrrrrrrrrrr}
\toprule
& & \multicolumn{4}{c}{Val-sin} & \multicolumn{4}{c}{Sim-k3} & \multicolumn{4}{c}{Lin-k3} & \multicolumn{4}{c}{Sim-ens} & \multicolumn{4}{c}{Lin-ens} \\
\cmidrule(lr){3-6}
\cmidrule(lr){7-10}
\cmidrule(lr){11-14}
\cmidrule(lr){15-18}
\cmidrule(lr){19-22}
Model & $H$ & Win & Med. & LQ & UQ & Win & Med. & LQ & UQ & Win & Med. & LQ & UQ & Win & Med. & LQ & UQ & Win & Med. & LQ & UQ \\
\midrule
Linear & 10 & 55.0 & 0.997 & 0.989 & 1.000 & 85.0 & 0.995 & 0.992 & 0.999 & 66.7 & 0.980 & 0.960 & 1.005 & 83.3 & 0.995 & 0.993 & 0.999 & 66.7 & 0.981 & 0.960 & 1.004 \\
 & 20 & 61.7 & 0.997 & 0.976 & 1.000 & 95.0 & 0.995 & 0.983 & 0.998 & 61.7 & 0.980 & 0.950 & 1.014 & 95.0 & 0.992 & 0.982 & 0.997 & 56.7 & 0.980 & 0.951 & 1.022 \\
 & 30 & 80.0 & 0.991 & 0.971 & 0.999 & 100.0 & 0.987 & 0.965 & 0.996 & 81.7 & 0.955 & 0.908 & 0.985 & 98.3 & 0.988 & 0.969 & 0.995 & 81.7 & 0.956 & 0.908 & 0.986 \\
\addlinespace[1pt]
MLP & 10 & 11.7 & 1.000 & 1.000 & 1.000 & 50.0 & 1.000 & 0.996 & 1.004 & 63.3 & 0.979 & 0.963 & 1.032 & 35.0 & 1.003 & 0.995 & 1.013 & 63.3 & 0.980 & 0.962 & 1.035 \\
 & 20 & 21.7 & 1.000 & 1.000 & 1.000 & 80.0 & 0.997 & 0.988 & 0.999 & 58.3 & 0.990 & 0.966 & 1.031 & 75.0 & 0.991 & 0.978 & 0.999 & 56.7 & 0.990 & 0.966 & 1.032 \\
 & 30 & 50.0 & 1.000 & 0.983 & 1.000 & 66.7 & 0.993 & 0.967 & 1.001 & 73.3 & 0.973 & 0.937 & 1.010 & 63.3 & 0.993 & 0.965 & 1.010 & 73.3 & 0.974 & 0.937 & 1.006 \\
\addlinespace[1pt]
LSTM & 10 & 40.0 & 1.000 & 0.978 & 1.013 & 56.7 & 0.997 & 0.978 & 1.012 & 56.7 & 0.991 & 0.958 & 1.042 & 61.7 & 0.994 & 0.982 & 1.012 & 53.3 & 0.995 & 0.959 & 1.042 \\
 & 20 & 36.7 & 1.000 & 0.992 & 1.018 & 50.0 & 1.000 & 0.987 & 1.017 & 50.0 & 1.000 & 0.976 & 1.038 & 48.3 & 1.001 & 0.988 & 1.019 & 50.0 & 1.000 & 0.975 & 1.039 \\
 & 30 & 30.0 & 1.000 & 0.997 & 1.025 & 40.0 & 1.002 & 0.992 & 1.017 & 46.7 & 1.003 & 0.982 & 1.033 & 38.3 & 1.005 & 0.993 & 1.014 & 48.3 & 1.000 & 0.981 & 1.036 \\
\addlinespace[1pt]
N-BEATS & 10 & 35.0 & 1.000 & 0.992 & 1.007 & 75.0 & 0.989 & 0.976 & 1.000 & 83.3 & 0.952 & 0.904 & 0.981 & 58.3 & 0.989 & 0.973 & 1.018 & 81.7 & 0.955 & 0.906 & 0.982 \\
 & 20 & 55.0 & 0.992 & 0.969 & 1.000 & 86.7 & 0.979 & 0.946 & 0.991 & 85.0 & 0.947 & 0.920 & 0.981 & 70.0 & 0.962 & 0.942 & 1.013 & 86.7 & 0.949 & 0.921 & 0.983 \\
 & 30 & 83.3 & 0.972 & 0.941 & 0.987 & 95.0 & 0.953 & 0.934 & 0.971 & 86.7 & 0.933 & 0.895 & 0.966 & 91.7 & 0.954 & 0.941 & 0.968 & 86.7 & 0.934 & 0.890 & 0.966 \\
\addlinespace[1pt]
PatchTST & 10 & 80.0 & 0.938 & 0.883 & 0.982 & 86.7 & 0.946 & 0.914 & 0.973 & 95.0 & 0.861 & 0.825 & 0.898 & 75.0 & 0.973 & 0.949 & 0.999 & 95.0 & 0.862 & 0.822 & 0.897 \\
 & 20 & 81.7 & 0.961 & 0.914 & 0.994 & 100.0 & 0.951 & 0.917 & 0.972 & 95.0 & 0.847 & 0.818 & 0.881 & 86.7 & 0.959 & 0.936 & 0.984 & 95.0 & 0.852 & 0.816 & 0.884 \\
 & 30 & 98.3 & 0.946 & 0.910 & 0.983 & 100.0 & 0.952 & 0.935 & 0.965 & 96.7 & 0.812 & 0.770 & 0.840 & 83.3 & 0.969 & 0.955 & 0.985 & 96.7 & 0.810 & 0.768 & 0.840 \\
\bottomrule
\end{tabular}
\caption{Horizon-wise version of Table~\ref{tab:fused_results}. For each model class and horizon, we report the win rate against default MIMO and the median, lower quartile (LQ), and upper quartile (UQ) of the test MSE ratio relative to default MIMO over asset--seed runs. Values below one indicate improvement.}
\label{tab:policy_by_model_by_horizon}
\end{table*}

\paragraph{Rollout block sizes.}
For each horizon, we use \(K=8\) candidate deployment block sizes. In implementation these are generated by evenly spacing values between 1 and \(H\), converting to integers, and removing duplicates:
\[
    \mathcal{S}_H
    =
    \mathrm{unique}\!\left(
    \left\lfloor
    \mathrm{linspace}(1,H,K)
    \right\rceil
    \right),
    \qquad K=8,
\]
where \(\lfloor \cdot \rceil\) denotes integer rounding as used in the implementation. This gives approximately uniform coverage from the fully recursive extreme \(s=1\) to the default MIMO rule \(s=H\), while keeping the candidate deployment pool small enough for validation-based policy selection.

For the reported horizons, this gives:
\[
\begin{aligned}
\mathcal{S}_{10} &= \{1,2,3,4,6,7,8,10\},\\
\mathcal{S}_{20} &= \{1,3,6,9,11,14,17,20\},\\
\mathcal{S}_{30} &= \{1,5,9,13,17,21,25,30\}.
\end{aligned}
\]

\paragraph{Data and assets.}
We use the VOLARE realized-variance stock archive and evaluate 20 univariate equity volatility series:
AAPL, ADBE, AMD, AMGN, AMZN, AXP, BA, CAT, CRM, CSCO, CVX, DIS, GE, GOOGL, GS, HD, HON, IBM, JNJ, and JPM.
For each asset, observations are sorted chronologically and the forecasting target is the log realized variance,
\[
    z_t = \log(\max(\mathrm{rv5\_ss}_t, 10^{-12})).
\]
Each series is split chronologically into 70\% train, 15\% validation, and 15\% test segments. Normalisation is fitted on the training segment only and then applied to validation and test:
\[
    \tilde z_t = \frac{z_t - \mu_{\mathrm{train}}}{\sigma_{\mathrm{train}} + 10^{-8}}.
\]
All models are trained and evaluated on the resulting normalized windowed datasets. Each input uses a window of length \(W=20\), and we evaluate horizons \(H \in \{10,20,30\}\).

\noindent \textbf{Model families.}
We evaluate five forecaster classes: LinearNet, MLPModel, LSTMModel, N-BEATS, and PatchTST. Each model is trained as a direct multi-output forecaster that predicts the full horizon \(H\) in one forward pass. All results use three random seeds, \(\{0,1,2\}\), for each asset, horizon, and model class.

\noindent \textbf{Hyperparameters.}
Unless otherwise stated, all models are trained with Adam, learning rate \(10^{-3}\), batch size 128, and 100 training epochs. The shared neural-network configuration uses embedding width 64, one layer, and 10\% dropout. PatchTST uses patch length 5, stride 2, and 4 attention heads. These values were selected using a short validation grid search over learning rate \(\{10^{-5},10^{-4},10^{-3}\}\), embedding width \(\{32,64,128\}\), and number of layers \(\{1,2,3\}\). Model selection is performed using validation rollout MSE over full horizon.

\noindent \textbf{Checkpoint selection.}
Training is run for the full epoch budget, but the final saved model restores the epoch with the best validation rollout MSE. For the induced-rule family, where a single \(H\)-output model can be rolled out with multiple block sizes, the selected checkpoint is the epoch with the best validation rollout MSE over the horizon. For the separately trained family, where one model is trained for each block size \(s\), the selected checkpoint is based on the validation rollout MSE for that same \(s\).

\section{Horizon Breakdown of Table 1}
\label{app:table1_app}

Table~\ref{tab:policy_by_model_by_horizon} provides a horizon-wise version of Table~\ref{tab:fused_results}, reporting results separately for \(H \in \{10,20,30\}\). The qualitative conclusions are consistent across horizons. In particular, non-default deployment policies continue to outperform default MIMO on a substantial fraction of runs, but the strongest policy remains dependent on both the model class and the horizon. 

Validation-selected singletons provide modest but low-cost improvements across all horizons, while subset and ensemble policies achieve larger median MSE reductions at higher inference cost. The relative performance of simple versus linear aggregation is also consistent with the main text: linear combinations often achieve stronger median MSE gains, but the advantage is not uniform across architectures or horizons. 

Overall, the horizon-wise breakdown confirms that the deployment headroom identified in Table~\ref{tab:policy_vs_mimo_grouped} is not driven by a particular forecast length, and that no single deployment policy dominates uniformly across horizons.






\end{document}